# Attention Is All You Need Until You Need Retention

## *Governing Persistent Transformer Memory with Social Learning Strategies*

**M. Murat Yaslioglu, Istanbul University, School of Business**

**Abstract**

Pretrained Transformers keep what they learned in their weights and lose what they observe once a session ends. The first version of this paper proposed a Retention Layer, a persistent memory that a Transformer block reads with attention and writes during use. Because most of what a deployed model could retain is produced by other agents, this revision treats deciding what to keep as a social learning problem: when to rely on observed behaviour, whom to learn from and how much independent agreement to require. We give a corrected specification of the layer, which reduces exactly to the base Transformer when its memory is empty. We derive the memory's lifecycle from social learning strategies: encoding gated by surprise, observed outcomes and earned credibility; consolidation by a credibility-weighted quorum of distinct, recent sources that must also outweigh every rival behaviour; and reconsolidation by the outcomes of reproduction. We prove that raising the quorum lowers the risk of consolidating a coordinated false template exponentially while delaying true templates only linearly, and that relative consolidation protects only while credible honest evidence arrives faster than adversarial evidence. In a simulation with world drift and three memory-poisoning attacks, the lifecycle reached accuracies of 0.989 to 0.996, against 0.62 to 0.63 for the ungated first-version design, and kept attack success at or below 0.07 when 30% of the observations about a target were adversarial. As predicted, it amplified attacks once adversarial evidence outpaced honest evidence. Experience with a long-running assistant adds two rules: a model's own outputs must not count as support, and a user's testimony should be kept after one mention. We close with an evaluation protocol for language models.

**Correspondence:** murat@yaslioglu.com; ORCID 0000-0003-2464-5439

## 1 Introduction

A pretrained Transformer (Vaswani et al. 2017) ends training with its knowledge fixed in its weights. At inference it conditions on a context window, and whatever it observes in one session is unavailable in the next unless something outside the model puts it back into the prompt. Context windows have grown from a few thousand tokens to hundreds of thousands and, in research systems, to millions (Gemini Team 2024). A longer window is still a window. Attending over it is expensive, its contents are used unevenly (Liu et al. 2024), and it is emptied when the session ends. Deployed assistants therefore attach memory at the system level, through retrieval over past interactions and managed memory stores (Lewis et al. 2020; Packer et al. 2023; Zhong et al. 2024; Chhikara et al. 2025), while a parallel line of research places memory inside the architecture (Y. Wu et al. 2022; Wang et al. 2024; Behrouz, Zhong, et al. 2025).

The first version of this paper (Yaslioglu 2025) proposed a Retention Layer: a persistent memory that a Transformer block reads with attention and writes during use, so that patterns observed in one session can shape behaviour in later ones. It motivated the design with the account of observational learning in social learning theory, in which learning from others depends on attention, retention, reproduction and motivation (Bandura 1986). The proposal has since been cited as an example of cognitively oriented designs for self-improving agents (Rahmani et al. 2026). It was also underspecified. The write operator was not defined, one superscript indexed both network depth and interaction time, one block equation dropped the self-attention output, training and evaluation were not discussed, and social learning theory served as an analogy rather than as a source of mechanisms.

This revision starts from an observation that makes the analogy do work. Almost everything a deployed model could retain is produced by someone else: a user who corrects it, a colleague who demonstrates a procedure, a document, a tool, another model. Deployment-time retention is observational learning from other agents, and it inherits the questions that social learning research asks about any such learner. When should observed behaviour override what the learner already believes? Whose behaviour is worth keeping? How much independent agreement should be required before a behaviour becomes durable? Cultural evolution and comparative psychology study these questions under the name of social learning strategies (Laland 2004; Rendell et al. 2010; Kendal et al. 2018). They describe heuristics such as copying when uncertain, copying successful or prestigious individuals and copying the majority, together with the conditions under which each heuristic fails.

The same questions have become practical engineering problems. Memory writes are an attack surface: adversaries can plant records that later steer an agent (Chen et al. 2024; Dong et al. 2025), and a growing set of defences scores trust, checks consensus or binds memories to their origin (Wei et al. 2025; Sunil et al. 2026; Louck 2026; Lin et al. 2026). Consolidation can also turn a hedged remark into a confident fact once its source is forgotten (Kwon 2026). In social learning terms these are failures to decide whom to copy and how much agreement to require.

The paper makes five contributions.

1. **A corrected specification of the Retention Layer** (Section 4). Reads use a null slot and a gate, depth and time are indexed separately, and a block whose memory is empty computes exactly the base Transformer block (Proposition 1). Writes are staged in an episodic buffer before they are consolidated into a bounded store.
2. **Lifecycle operators derived from social learning strategies** (Sections 2.2, 4.4 and 4.5). Encoding is gated by surprise (copy when uncertain), by observed outcomes (payoff bias) and by source credibility earned through past success (prestige bias). Consolidation follows conformist transmission: a behaviour is adopted only when the credibility-weighted support of distinct, recent sources reaches a quorum that grows with its conflict with the model's current belief, and exceeds the support for every rival behaviour, including the one the model already produces. Retrieval is weighted by retention strength, which confirmations and the outcomes of reproduction revise and disuse erodes.
3. **An analysis of quorum consolidation** (Section 5). Under stated assumptions, the probability of consolidating a coordinated false template within a window decays exponentially in the quorum while the expected delay before a true template is consolidated grows linearly. With relative consolidation, a false template is ever consolidated with a probability that vanishes exponentially in the quorum when credible honest evidence arrives faster, and with certainty when it does not, so conformity protects only while honest evidence outpaces the adversary's.
4. **A mechanism-level proof of concept** (Section 6). A controlled simulation with world drift and three attack models compares the lifecycle with the first version's ungated design and with surprise-only and trust-only write policies, and ablations isolate each component.
5. **An evaluation protocol and lessons from deployment** (Sections 7 to 10). Section 7 proposes experiments for language-model instantiations. Section 8 discusses applications and reports lessons from a long-running single-user assistant, which add requirements that the simulation could not reveal: a model's own outputs must not count as support, a user's testimony about themselves should be kept after one mention, and the read path must be monitored. Sections 9 and 10 cover limitations and governance.

The paper remains conceptual and analytical. We do not train a language model with a Retention Layer, and the simulation abstracts token sequences into vectors. Its purpose is to check that the specified operators behave as claimed and to expose the trade-offs that a full-scale study will have to measure.

## 2 Background

### 2.1 Observational learning

Social learning theory treats imitation as learning from models, meaning people whose behaviour is observed. In Bandura's account, observers do not copy responses mechanically. They transform modelled behaviour into symbolic representations that later guide performance, and they refine performance by comparing it with those representations (Bandura 1986). Experiments supported both parts of this account. Symbolic coding and rehearsal of modelled actions improved their later reproduction (Bandura and Jeffery 1973), and the accuracy of the observer's representation carried the effect of repeated viewing and verbal coding on the accuracy of reproduction (Carroll and Bandura 1990). Whether learned behaviour is performed depends on expected outcomes, including outcomes observed to befall the model, which is usually called vicarious reinforcement (Bandura 1986).

Summaries of this account often list four processes: attention to the model, retention of a coded representation, reproduction of the behaviour and motivation to perform it. The first version of this paper used that summary, and fuller treatments are available elsewhere (Bandura 1986; Schunk 2012). We keep the four processes with one qualification: they interact rather than run in sequence. Motivation, for example, shapes what is attended to and rehearsed, not only what is performed.

### 2.2 Social learning strategies

Learning from others is cheap but unreliable, because demonstrators may be wrong, outdated or deceptive. Social learning strategies are the rules that determine when individuals use social information and from whom (Laland 2004). Laland distinguished *when* strategies, such as copying when individual learning is costly, when uncertain or when dissatisfied with one's own results, from *who* strategies, such as copying the majority, copying successful individuals or copying when the demonstrator does better. A computer tournament of strategies found that heavy reliance on copying could win, largely because demonstrators tended to display behaviour that had already worked for them, which made copying an efficient filter (Rendell et al. 2010). Prestige bias offers a shortcut to copying the successful, since deference accumulates to individuals whose skill others have observed (Henrich and Gil-White 2001). Conformist transmission, a disproportionate tendency to adopt the most common behaviour, raises the reliability of what is adopted when most demonstrators are right, and it can also stabilise behaviour that is common but wrong (Boyd and Richerson 1985). Human experiments show that people use social information more when uncertain and that conformity depends on the number of demonstrators (Morgan et al. 2012; Muthukrishna et al. 2016). Kendal et al. (2018) review how these strategies combine and how their use varies across species and contexts.

Two features of this literature matter for machine memory. First, the strategies are defined by the *source* of information and by *agreement across sources*, two quantities that memory designs for language models seldom represent. Second, every strategy has a characteristic failure. Copying when uncertain is exploitable wherever the learner knows least; copying the successful is exploitable when outcomes can be faked; prestige can be counterfeited; and conformity amplifies coordinated majorities. A memory policy built on these strategies should inherit these failure modes, and they become testable predictions.

### 2.3 What pretrained Transformers retain

The first version stated that context windows span 2,000 to 4,000 tokens. That figure was outdated even in January 2025, when commercial models accepted hundreds of thousands of tokens and research systems had reported millions (Gemini Team 2024). The underlying point needs a sharper statement. At inference a standard Transformer draws on two sources of information: parameters fixed by pretraining and fine-tuning (Brown et al. 2020; OpenAI 2023), and the current context. Within a context, attention can implement learning-like computations, including steps that resemble gradient descent on in-context examples (Oswald et al. 2023; Akyürek et al. 2023). These effects vanish with the context. Parameters can be changed after deployment through dynamic evaluation (Krause et al. 2018), knowledge editing (Meng et al. 2022) or test-time training (Sun et al. 2024), but such changes require optimisation steps and can interfere with existing knowledge (Kirkpatrick et al. 2017). What a standard Transformer lacks is a mechanism that carries observed information into future sessions without a training step and that decides, observation by observation, what deserves to be kept.

## 3 Related work

### 3.1 External and differentiable memory

Memory networks read from a set of stored vectors by content-based attention (Weston et al. 2015; Sukhbaatar et al. 2015). Neural Turing Machines and the Differentiable Neural Computer added learned writes with erase and add vectors, addressed by content and location (Graves et al. 2014, 2016), and networks with such memories trained on episodic curricula learned new classes from single examples (Santoro et al. 2016). The attention-based write option in the first version of this paper belongs to this family.

### 3.2 Memory-augmented Transformers

Several architectures carry state across segments or documents. Memformer maintains an external memory with separate reader and writer modules (Q. Wu et al. 2022); the Recurrent Memory Transformer passes memory tokens between segments (Bulatov et al. 2022); Token Turing Machines read and write a small set of memory tokens through token summarisation (Ryoo et al. 2023); and TransformerFAM feeds a layer's own activations back as working memory (Hwang et al. 2024). Memorizing Transformers retrieve from a non-differentiable cache of past key-value pairs by approximate nearest-neighbour search (Y. Wu et al. 2022), in the spirit of nearest-neighbour language models (Khandelwal et al. 2020) and retrieval-enhanced models (Borgeaud et al. 2022; Lewis et al. 2020). LongMem decouples a frozen memory encoder from a trainable reader (Wang et al. 2023). MemoryLLM embeds a fixed-size pool of memory tokens in every layer and updates it as new knowledge arrives (Wang et al. 2024), and M+ adds a long-term store with a co-trained retriever (Wang et al. 2025). LM2 augments a decoder with a memory module accessed through gated cross-attention (Kang et al. 2025). Parametric key-value memory layers (Lample et al. 2019; Berges et al. 2024) can be updated sparsely to add knowledge with less forgetting (Lin et al. 2025). Omidi et al. (2025) review this area from a neuroscience perspective and single out write policies and consolidation as unresolved problems.

### 3.3 Test-time memorisation and fast weights

Fast-weight models write associations into rapidly changing weights (Schmidhuber 1992; Ba et al. 2016), and linear attention can be read as a fast-weight programmer (Schlag et al. 2021). Test-time training layers update an inner model by gradient steps on a self-supervised loss (Sun et al. 2024). Titans adds a neural long-term memory that is updated at test time by a surprise signal with momentum and an adaptive forgetting gate (Behrouz, Zhong, et al. 2025). MIRAS recasts sequence models as associative memories optimised online, defined by an attentional-bias objective and a retention regulariser (Behrouz, Razaviyayn, et al. 2025a); ATLAS optimises its memory over a window of past inputs rather than the current token alone (Behrouz, Li, et al. 2025); and Nested Learning organises model components into a continuum of memories that update at different frequencies (Behrouz, Razaviyayn, et al. 2025b). These models mainly target long sequences, and their write signals depend on the input rather than on who produced it.

A note on terminology. RetNet uses *retention* for a recurrent mechanism with exponential decay that replaces attention (Sun et al. 2023), and MIRAS uses the word for the regulariser that controls forgetting. In this paper *retention* names the lifecycle that decides what persists across sessions: encoding, consolidation, retrieval, reconsolidation and forgetting.

### 3.4 Cognitively inspired memory for language agents

Agent frameworks treat memory as a system component. Generative Agents retrieve from a memory stream by recency, importance and relevance and write higher-level reflections (Park et al. 2023). MemoryBank applies a forgetting curve with strengthening on recall (Zhong et al. 2024), related work models human-like recall and consolidation for personal agents (Hou et al. 2024), MemGPT manages a hierarchy of context and external storage (Packer et al. 2023), and CoALA organises agent memory into working, episodic, semantic and procedural stores (Sumers et al. 2024). Neuroscience has inspired event segmentation of token streams (Fountas et al. 2025), episodic memory control grounded in complementary learning systems (Das et al. 2024; McClelland et al. 1995; Kumaran et al. 2016), training-free consolidated associative memory (He et al. 2024) and hippocampal indexing for retrieval (Gutiérrez et al. 2024). Recent systems link notes that evolve over time (Xu et al. 2025), extract and consolidate salient facts (Chhikara et al. 2025), consolidate only when similar interactions recur (Dai et al. 2026), or train consolidation against a verifier of coverage, preservation and faithfulness (Yang et al. 2026). Surveys cover the rapidly growing space (Zhang et al. 2025; Hu et al. 2025). With few exceptions, these designs model memory as individual cognition. They decide what to keep from the content and statistics of observations, not from the identity and track record of whoever produced them.

### 3.5 Memory security and governance

Poisoning attacks optimise triggers that retrieve planted records (Chen et al. 2024) or inject malicious records through ordinary queries (Dong et al. 2025). Defences include consensus-based validation across related memories combined with a separate store of lessons from detected failures (Wei et al. 2025), composite trust scores and trust-aware retrieval with temporal decay (Sunil et al. 2026), Bayesian trust over memory sources (Bhardwaj 2026), and origin-bound authority that requires independent trusted corroboration before untrusted memory can trigger consequential actions (Louck 2026). Governance frameworks add consistency checks before consolidation, temporal decay and access control (Lam et al. 2026), and a recent survey organises attacks and defences along the memory lifecycle (Lin et al. 2026). Kwon (2026) shows that consolidation can erase the difference between hedged and asserted claims and that a single redundant source can restore correct decisions. The present proposal gives these mechanisms a common rationale, predicts when each of them fails, and moves them into the lifecycle of an architectural memory.

### 3.6 Social learning in artificial agents

Reinforcement learning agents can learn to exploit cues from expert agents that share their environment (Ndousse et al. 2021), and agents trained in rich 3D worlds can imitate a demonstrator in real time and keep the behaviour after the demonstrator leaves, a form of cultural transmission (Bhoopchand et al. 2023). Language models can teach one another in natural language while limiting the leakage of private data (Mohtashami et al. 2023). Hillier et al. (2024) survey social learning in agents more broadly. To our knowledge, social learning strategies have not previously been used to govern the persistent memory of a Transformer. Table 1 positions the proposal against representative designs.

Table 1. Positioning against representative memory designs. "Source-aware" means that the identity or track record of the producer of an observation affects what is written or kept.

| Design | Persists across sessions | Inside the architecture | Write or consolidation signal | Source-aware |
|---|---|---|---|---|
| Memorizing Transformers (Y. Wu et al. 2022) | Not by design | Yes | Recent key-value pairs are cached | No |
| MemoryLLM, M+ (Wang et al. 2024, 2025) | Yes | Yes | Injection of new knowledge | No |
| Titans (Behrouz, Zhong, et al. 2025) | Long sequences | Yes | Surprise, with momentum and a forgetting gate | No |
| Generative Agents (Park et al. 2023) | Yes | No | All observations; retrieval by recency, importance, relevance | No |
| MemoryBank (Zhong et al. 2024) | Yes | No | All interactions; strengthening on recall | No |
| RecMem (Dai et al. 2026) | Yes | No | Recurrence of similar interactions | No |
| Trust-scored agent memory (Sunil et al. 2026; Bhardwaj 2026) | Yes | No | Trust scores | Partly |
| Retention Layer, first version (Yaslioglu 2025) | Intended | Yes | Unspecified (append, or conditional on feedback) | No |
| This paper | Yes | Yes | Surprise, outcomes, credibility; relative quorum scaled by conflict | Yes |

# 4 The Retention Layer

## 4.1 Design goals and notation

We index interaction sessions by $t$ and Transformer blocks by $\ell \in \{1, \dots, L\}$. A session is any unit after which the context is discarded, such as a conversation, a task episode or a document. $X_t^{(\ell)} \in \mathbb{R}^{n \times d}$ denotes the residual stream entering block $\ell$ during session $t$. A chosen subset of blocks carries a retention store

$$M_t^{(\ell)} = \{(k_i, v_i, s_i, S_i)\}_{i=1}^{m},$$

in which $k_i \in \mathbb{R}^{d_m}$ is a key describing a situation, $v_i \in \mathbb{R}^{d_m}$ is a value describing a behaviour template, $s_i > 0$ is a retention strength and $S_i$ is the set of sources whose observations support the entry. Each such block also keeps an episodic buffer $B_t^{(\ell)}$ of candidate entries that await consolidation. The first version borrowed the term from Baddeley's model of working memory (Baddeley 2000); here it simply denotes a short-term staging store.

The specification pursues five goals, each of which answers a gap in the first version:

- **G1.** Inserting the layer must not change a pretrained model until memory holds something relevant.
- **G2.** The read must be able to abstain.
- **G3.** Network depth and interaction time must be indexed separately.
- **G4.** Memory size and read cost must stay bounded.
- **G5.** Writes must be selective and aware of their sources.

### 4.2 Block equations and the read operator

We use the pre-normalisation block. A memory-bearing block computes

$$Z_t^{(\ell)} \& = X_t^{(\ell)} + MHA\left(LN_1(X_t^{(\ell)})\right), \\ Y_t^{(\ell)} \& = Z_t^{(\ell)} + Read\left(LN_2(Z_t^{(\ell)}); M_{t-1}^{(\ell)}\right), (1) \\ X_t^{(\ell+1)} \& = Y_t^{(\ell)} + FFN\left(LN_3(Y_t^{(\ell)})\right).$$

The store read during session $t$ is the one produced at the end of session $t-1$. Memory is not passed from layer to layer. It changes between sessions through the lifecycle of Section 4.5,

$$\left(M_t^{(\ell)}, B_t^{(\ell)}\right) = Lifecycle\left(M_{t-1}^{(\ell)}, B_{t-1}^{(\ell)}, O_t, F_t\right), (2)$$

where $O_t$ is the set of observations made during the session and $F_t$ is the feedback received on the model's own outputs. The first version wrote $R^{(l)}, M^{(l+1)} = Retention(X^{(l)}, M^{(l)})$ and described $M^{(l+1)}$ as the state seen by "the next layer (or next inference step)", which conflates the two indices (G3). Its Section 3 also set $X^{(l+1)} = X^{(l)} + R^{(l)} + dropout(O^{(l)})$ with $O^{(l)} = FFN(X^{(l)} + R^{(l)})$, so the self-attention output never reached the residual stream. Equation (1) corrects both problems.

For a token state $h$, that is, a row of $LN_2(Z_t^{(\ell)})$, the read is

$$a_0 \& = b_0, a_i = \frac{(W_Q h)^\top W_K k_i}{\sqrt{d_k}} + \beta \log s_i (i=1,\ldots,m), \\ (\alpha_0,\ldots,\alpha_m) \& = softmax(a_0,\ldots,a_m), \\ Read(h;M) \& = \sigma\left(w_g^\top h + b_g\right) W_O \sum_{i=1}^{m} \alpha_i W_V v_i. \quad (3)$$

Index 0 is a null slot with a learned logit $b_0$ and a zero value. When no stored key matches, attention flows to the null slot and the read returns approximately zero (G2). Without a null slot, as in the first version, the softmax must spread all of its mass over stored entries and the read injects a mixture of irrelevant templates. The term $\beta \log s_i$ multiplies the unnormalised attention weight of an entry by $s_i^\beta$, so strong entries win ties between similar keys. The sum of a context-match term and a strength term has the same form as activation-based retrieval in the ACT-R cognitive architecture, where the activation of a memory chunk adds a base level that reflects the frequency and recency of its use to its association with the current context (Anderson et al. 2004). A multi-head read applies Equation (3) per head with separate projections. For large stores the sum can be restricted to the top-$k$ keys returned by approximate nearest-neighbour search, as in Memorizing Transformers (Y. Wu et al. 2022).

**Proposition 1 (Reduction).** If the store is empty ($m=0$), or $W_O=0$, or the gate is identically zero, a memory-bearing block computes exactly the base pre-normalisation block.

*Proof.* In each case $Read(\cdot)=0$ for every token, so $Y_t^{(\ell)}=Z_t^{(\ell)}$ and Equation (1) reduces to the base block. ▫

Proposition 1 secures G1. With $W_O$ initialised at zero, the layer can be inserted into a pretrained model without changing its outputs, the same device used by zero-initialised gated cross-attention in Flamingo (Alayrac et al. 2022) and zero-initialised attention in LLaMA-Adapter (Zhang et al. 2023). The released code checks all three cases numerically on a random block; the maximum absolute difference from the base block is exactly zero in each case, and it is non-zero for a non-empty store with a non-zero output projection.

### 4.3 Observations, sources and outcomes

An observation $o=(c,b,\sigma,y)$ consists of a situation span $c$ and a behaviour span $b$ in the context of a session, a source identifier $\sigma$ and an outcome $y\in\{-1,+1,\varnothing\}$, where $\varnothing$ means that no outcome was observed. In a coding assistant, $c$ might be a request, $b$ the fix that a user wrote after rejecting the model's attempt, $\sigma$ the user's account and $y$ whether the tests then passed. Sources can be authenticated users, services, documents identified by origin, or other agents. Outcomes can come from explicit ratings, task success signals or later corrections. When the spans cannot be identified, $c$ and $b$ can both be taken as the whole observation.

The first version encoded a write vector as the mean of the block's token states, which discards order and mixes the situation with the behaviour. We encode keys and values by attention pooling with learned seed queries (Lee et al. 2019):

$$k(o)=\sum_{j\in c} softmax_j\left(\frac{p_K^\top U_K h_j}{\sqrt{d_m}}\right)U_V h_j,(4)$$

and $v(o)$ is defined in the same way over the span $b$ with its own seed query and projections. The first version also proposed storing templates symbolically, for example as parse trees or knowledge-graph triples. Nothing in the lifecycle requires values to be vectors: a value can be a structured template, such as a tool-call sequence, that is retrieved through its vector key and decoded into the context, which keeps stored behaviour open to inspection.

### 4.4 Social learning signals

Each observation receives five signals, all computed from quantities that the model already has or can maintain cheaply.

*Surprise (copy when uncertain).* The length-normalised negative log-likelihood of the observed behaviour under the current model with its current memory is

$$\upsilon(o)=-\frac{1}{|b|}\sum_{j\in b}\log p_\theta\left(x_j / x_{¿j}; M_{t-1}\right).(5)$$

It is high when the model is uncertain and also when the model confidently expects something else. When the demonstration is part of the context, Equation (5) comes from the forward pass that processes it. Titans uses a gradient-based measure of surprise for a similar purpose (Behrouz, Zhong, et al. 2025).

*Payoff (vicarious reinforcement).* $\pi(o)=1$ if $y=+1$, $\pi(o)=0$ if $y=-1$, and $\pi(o)=1/2$ if no outcome was observed.

*Credibility (prestige earned by success).* Each source keeps Beta pseudo-counts $(a_\sigma, b_\sigma)$ and has credibility $\chi_\sigma = a_\sigma / (a_\sigma + b_\sigma)$. The counts grow when the source's demonstrations are observed to succeed or fail, and when entries that the source supports lead to successful or failed reproductions (Equation 8). New identities start from a weak prior such as $Beta(1,1)$, and an account's history can supply an informative prior. Credibility can also be kept separately for different domains.

*Conflict.* The strength with which the model's current belief opposes the observed behaviour is

$$\Delta(o) = max\left(0, \frac{1}{|\hat{b}|} \log p_\theta(\hat{b}/c) - \frac{1}{|b|} \log p_\theta(b/c)\right), (6)$$

where $\hat{b}$ is the model's own continuation for $c$. Novel situations give $\Delta \approx 0$, and observations that contradict confident knowledge give large values.

*Support (conformity).* Every buffer entry and stored slot keeps the set $S$ of distinct sources that have supported it, each with the last time $t_\sigma$ at which it did so. The support of an entry $e$ at time $t$ is the credibility-weighted number of sources seen within a recency window $W$,

$$Q_W(e) = \sum_{\sigma \in S_e : t - t_\sigma \le W} \chi_\sigma .$$

Because support counts distinct sources rather than observations, repetition by a single source adds nothing; without this property, consolidation can turn repeated hearsay into fact (Kwon 2026). The window makes the count describe the current population of demonstrators, which is what conformist transmission responds to, and it lets legitimate change displace old support.

### 4.5 Lifecycle operators

**Encoding (retention, first stage).** Each observation is first compared with what the store and the model already hold.

- *Confirmation.* If the observed behaviour matches a stored slot $i$, meaning that its key and value lie within cosine similarity $\delta$, the source is recorded in $S_i$ with the current time, and $s_i$ grows by $\chi_\sigma$ if $\sigma$ is new to the slot. An observed failure ($y = -1$) instead multiplies $s_i$ by $e^{-\eta}$.
- *Tally.* If no slot matches but the model already produces the behaviour, that is, $\upsilon(o) \le \upsilon_0$, the source is tallied in a buffer entry for that behaviour. Tallies cost almost nothing and are never consolidated by themselves; they record how much credible support the model's current behaviour enjoys.
- *Candidate.* Otherwise the observation is novel or contradicts the model, and it enters the buffer only if

$$\theta_\upsilon(\upsilon(o) - \upsilon_0) + \theta_\pi(\pi(o) - 1/2) + \theta_\chi(\chi_\sigma - 1/2) > 0 . (7a)$$

A candidate merges into an existing entry for the same situation and behaviour, which adds $\sigma$ to $S_e$ and updates the outcome counts $n_e^+$ and $n_e^-$; a tally that receives a candidate becomes a candidate entry. Otherwise the candidate opens a new entry. When the buffer is full, the least recently updated entry is dropped.

**Consolidation (retention, second stage).** A candidate entry $e$ for situation $c$ moves to the store when

$$Q_W(e) \ge k_0 + \gamma \Delta(e), n_e^+ \ge n_e^-, Q_W(e) > \max_{e' \in R(e)} Q_W(e'), (7b)$$

where $R(e)$ collects the buffer entries, tallies and stored slots that hold a different behaviour for the same situation. The first condition is an absolute quorum. With $\gamma=0$ it is fixed; with $\gamma>0$ the evidence required grows with the conflict, so the model learns quickly where it knows nothing and demands more agreement before it overturns what it believes. The second condition is payoff bias. The third makes consolidation relative: a behaviour is adopted only if it currently has more credible support than every rival, including the behaviour the model already produces. Without it, a surprise-gated memory ignores confirmations of what the model already does while admitting every contradiction, so a persistent adversary accumulates support that honest sources are never credited with (Section 6.2). The consolidated entry merges into a matching slot, adding the credibility of sources that are new to the slot, or is inserted with $s=Q_W(e)$. Slots with similar keys and different values may coexist, and retrieval arbitrates between them through their strengths. If the store is full, the weakest slot is evicted provided it is weaker than the incoming entry.

**Reproduction and reconsolidation (motivation).** When the model produces $\hat{b}$ and receives feedback $y \in \{-1, +1\}$, let $\acute{\alpha}_i$ be the gated read weight of slot $i$ averaged over the tokens of $\hat{b}$ and over heads. Then

$$s_i \leftarrow s_i \exp(\eta \acute{\alpha}_i \psi_i), \psi_i = \begin{cases} y & \text{if } v_i \text{ agrees with } \hat{b}, \\ -1 & \text{if } v_i \text{ disagrees with } \hat{b} \text{ and } y=+1, \\ 0 & \text{otherwise}, \end{cases} \quad (8)$$

and every supporter $\sigma \in S_i$ has $\acute{\alpha}_i |\psi_i| / |S_i|$ added to $a_\sigma$ if $\psi_i > 0$ or to $b_\sigma$ if $\psi_i < 0$. Agreement can be measured by the similarity between $v_i$ and the pooled encoding of $\hat{b}$. Entries that produced a successful behaviour are strengthened and entries that produced a failure are weakened. After a success, entries that pointed elsewhere are weakened as well. After a failure, entries that disagreed are left unchanged, because the feedback does not say which alternative was right. The rule echoes two findings about human memory without depending on them: retrieval practice strengthens retention (Roediger and Butler 2011), and retrieved memories can be modified before they are stabilised again (Nader et al. 2000).

**Forgetting.** Strengths decay between sessions, $s_i \leftarrow (1-\lambda) s_i$. Power-law decay, which matches the statistics of how often information is needed again (Anderson and Schooler 1991), is a drop-in alternative.

Algorithm 1 collects the operators, and Table 2 maps them to the processes of observational learning and to social learning strategies.

```
Algorithm 1. Retention lifecycle for one memory-bearing block, applied after session t
Input: store M, buffer B, observations O_t, feedback F_t on the model's outputs
for each (output b^, outcome y) in F_t:                      # motivation
    for each slot i read while producing b^:
        s_i <- s_i * exp(eta * abar_i * psi_i)               # Eq. 8
        update (a_s, b_s) for every source s in S_i
for each observation o = (c, b, s, y) in O_t:
    if y is observed: update (a_s, b_s) with y               # vicarious outcome
    if b matches a slot i of M:                              # confirmation
        if y = -1: s_i <- s_i * exp(-eta)
        else: record s in S_i at time t; if s is new, s_i += chi_s
    else if surprise(o) <= upsilon_0:                        # tally
        record s at time t in the tally entry for (c, b) in B
    else if admit(o):                                        # Eq. 7a
        e <- merge o into its entry of B, or open a new entry
        if Q_W(e) >= k0 + gamma * Delta(e) and n+_e >= n-_e
                and Q_W(e) > Q_W(r) for every rival r of e:  # Eq. 7b
            merge e into M, or insert it with strength Q_W(e)
```

```
            if M is over capacity: evict the weakest slot
            remove e from B
for each slot i of M: s_i <- (1 - lambda) * s_i              # forgetting
```

Table 2. Correspondence between observational learning, social learning strategies and the lifecycle.

| Observational learning process | Social learning strategy | Operator | Signal |
|---|---|---|---|
| Attention | Copy when uncertain | Encoding gate (7a); read (3) | Surprise $\upsilon$; attention weights |
| Retention | Prestige bias; conformist transmission | Confirmations, tallies and relative quorum consolidation (7b) | Credibility $\chi_\sigma$; windowed support $Q_W$; conflict $\Delta$ |
| Reproduction | Performance guided by a stored representation | Strength-weighted read (3) | Strength $s_i$ |
| Motivation | Payoff bias; vicarious reinforcement | Reconsolidation (8); credibility update | Outcomes $y$; payoff $\pi$ |

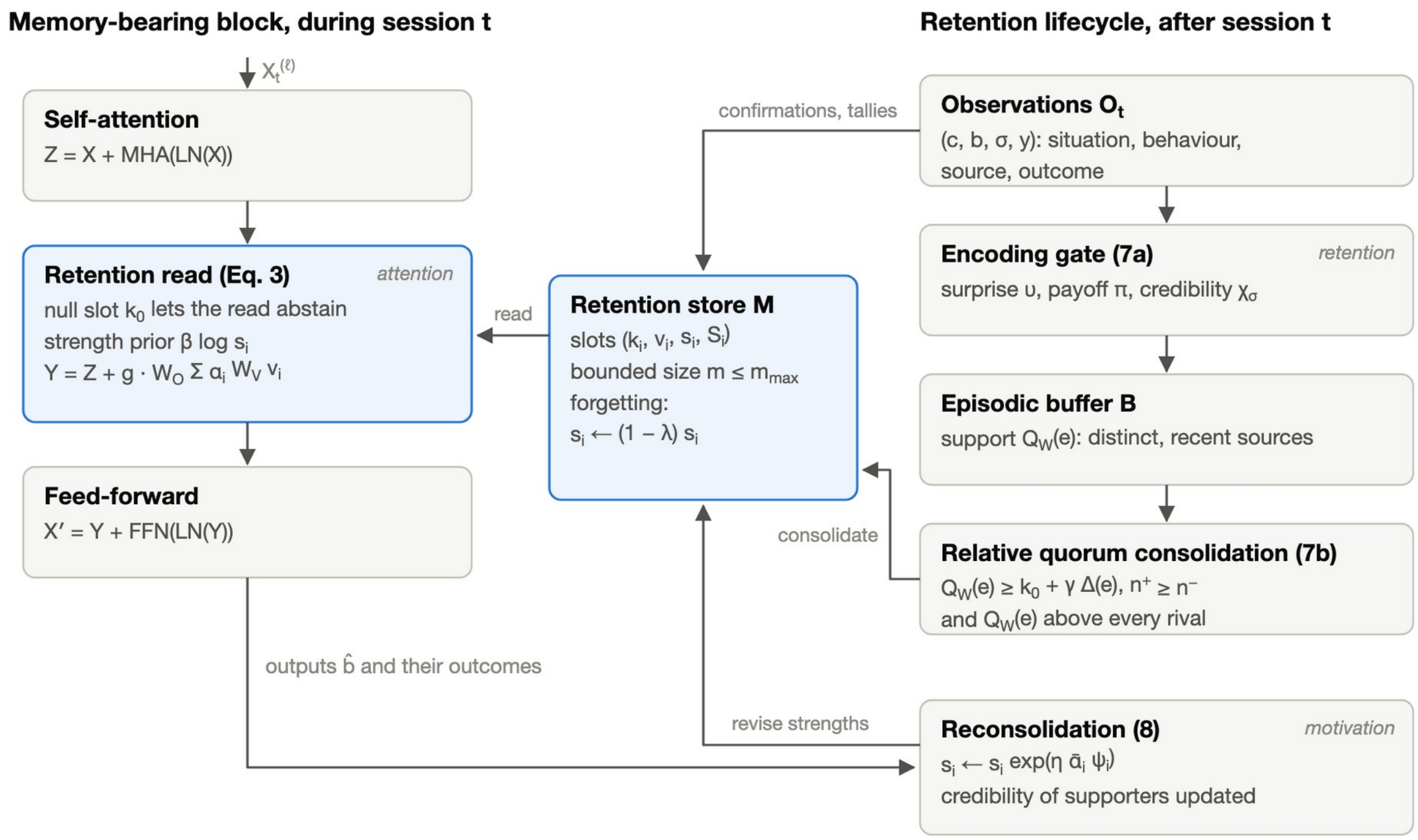

Figure 1. A memory-bearing Transformer block and the retention lifecycle. During a session the block reads the store through gated attention with a null slot. After the session, observations that confirm a stored slot add support to it, observations of behaviour the model already produces are tallied, and other observations that pass the encoding gate enter the episodic buffer. An entry is consolidated when the credibility-weighted support of distinct, recent sources reaches the quorum and exceeds that of every rival behaviour. Feedback on the model's outputs reconsolidates the slots that produced it and updates source credibility, and strengths decay.

## 4.6 Training

The lifecycle is not differentiable end to end, and it does not need to be. Three groups of parameters can be treated separately.

*Read path and encoders.* The projections and gate in Equation (3) and the pooling parameters in Equation (4) can be trained on multi-session episodes. Earlier sessions are run and the store is built with the lifecycle; next-token loss is then minimised on later sessions that require the stored behaviour, with store contents treated as constants at session boundaries. Memorizing Transformers train reads against a non-differentiable memory in the same way (Y. Wu et al. 2022). Gradients reach the encoders through entries written within the same training episode. Mixing in episodes where memory is irrelevant or adversarial teaches the null slot and the gate to abstain.

*Lifecycle parameters.* The gate weights $\theta_\upsilon, \theta_\pi, \theta_\chi$, the reference surprise $\upsilon_0$, the quorum parameters $k_0$ and $\gamma$, the step size $\eta$ and the decay $\lambda$ are few and interpretable. They can be tuned on validation episodes or fitted by policy-gradient methods that reward a write by its effect on later loss.

*Base weights.* The base weights can stay frozen, or stored entries can be re-encoded after any update. Keys and values produced by old weights become stale when the model changes, a problem shared by every latent memory; keeping the source spans of consolidated entries makes re-encoding possible.

### 4.7 Cost

A dense read costs $O(n(m+1)d_k)$ per memory-bearing block, and $O(n(k\,d_k+\log m))$ with approximate top-$k$ retrieval. The store holds at most $m_{max}$ entries and the buffer at most $B$, so the state is bounded however many sessions occur (G4). Surprise and conflict require the likelihood of the observed behaviour and of the model's own continuation. The first is already computed when the behaviour is in the context, and the second can reuse the output that the model produced during the session. Consolidation and reconsolidation touch only matching entries and the slots read while an output was produced.

## 5 Analysis of quorum consolidation

Consider one situation and the stream of observations about it. We adopt assumptions that favour the adversary.

- **A1.** Each observation is adversarial with probability $\rho$, independently of the others. All adversarial observations show the same false template $a$ and come from distinct identities with credibility $\chi_a$.
- **A2.** Otherwise the observation comes from a distinct honest source with credibility $\chi_h$ and shows the true template $b$ with probability $q$, or an idiosyncratic error that no other source repeats.
- **A3.** Every observation passes the encoding gate, outcomes are ignored, credibilities are constant and consolidation uses a fixed quorum $k$ (that is, $\gamma=0$).

Let $K_a=\lceil k/\chi_a \rceil$ and $K_h=\lceil k/\chi_h \rceil$ be the numbers of adversarial and honest-correct observations needed to reach the quorum.

**Proposition 2 (Quorum trade-off).** Under A1 to A3:

(a) The probability that $a$ is consolidated within the first $N$ observations is $P_{adv}(N)=Pr[Bin(N,\rho)\geq K_a]$. If $K_a>N\rho$, then $P_{adv}(N)\leq \exp(-N\,KL(K_a/N \,||\, \rho))$, where $KL$ denotes the Kullback-Leibler divergence between Bernoulli distributions.

(b) The expected number of observations until $b$ is consolidated is $E[T_b]=K_h/((1-\rho)q)$.

(c) Let $T_a$ and $T_b$ be the numbers of observations until the support of $a$ and of $b$ first reaches $k$. As $k\to\infty$, $Pr[T_a<T_b]\to 1$ if $\chi_a\rho>\chi_h(1-\rho)q$, and $Pr[T_a<T_b]\to 0$ if $\chi_a\rho<\chi_h(1-\rho)q$.

(d) Suppose consolidation also requires the relative condition of Equation (7b), with an unbounded window. If $\chi_a \rho < \chi_h (1-\rho) q$, the probability that $a$ is ever consolidated tends to zero exponentially fast as $k \to \infty$. If $\chi_a \rho > \chi_h (1-\rho) q$, $a$ is consolidated with probability one.

*Proof.* (a) After $N$ observations the adversarial support is $\chi_a$ times a $Bin(N, \rho)$ count, and it reaches $k$ exactly when the count reaches $K_a$. The bound is the Chernoff bound for the upper tail of a binomial distribution. (b) Honest-correct observations form a Bernoulli process with success probability $(1-\rho)q$, and $T_b$ is the waiting time for its $K_h$-th success, a negative binomial variable with mean $K_h/((1-\rho)q)$. (c) $T_a$ and $T_b$ are negative binomial with means that grow linearly in $k$ and standard deviations that grow as $\sqrt{k}$; when the two credibility-weighted rates differ, the gap between the means grows linearly in $k$, and Chebyshev's inequality around the midpoint gives the limit. (d) The difference between adversarial and honest support is a random walk. With a higher honest rate it drifts downward and lies far below zero when the adversarial support first reaches $k$, and Lundberg's inequality bounds the probability that it ever climbs back above zero. Appendix A gives the details. ▫

The proposition says three things. Raising the quorum increases the delay before a true template is consolidated in proportion to $k$ (part b), while the chance that a coordinated false template gets in within any fixed window falls exponentially once $k$ exceeds $\chi_a N \rho$ (part a). Credibility weighting multiplies the number of identities an attacker needs by $1/\chi_a$, so fresh identities with weak priors cost more than established ones; an attacker who farms reputation removes that factor. Finally, conformity cuts both ways (part c). A large quorum amplifies whichever side has the larger credibility-weighted arrival rate, just as conformist transmission can entrench a common error (Boyd and Richerson 1985). Quorums protect only when honest, credible evidence arrives faster than the adversary's.

Part (d) shows what the relative condition adds. With an absolute quorum alone, both templates are eventually consolidated whatever their rates, and only retrieval strengths separate them afterwards. With the relative condition, the probability that a coordinated false template is ever consolidated falls exponentially in the quorum while credible honest evidence arrives faster, and the false template wins for certain when it does not. The quorum then decides how quickly the contest is settled and how much it costs in delay.

A Monte Carlo check with $N=20$, $q=0.8$, $\chi_a=0.5$ and $\chi_h=0.8$, four values of $\rho$ and seven quorums (200,000 trials each) matches the exact expression in part (a) to within 0.0023 in absolute probability, and the simulated delays agree with part (b) to within 2.2%. For part (d), 5,000 simulated streams of 1,500 observations with $\rho=0.3$ give probabilities of 0.32, 0.16, 0.045, 0.0056 and 0 that the false template is ever consolidated under the relative condition for $k=1,2,4,8$ and $16$. With $\rho=0.6$, where the adversarial rate is the higher one, the probability is at least 0.9998 for every quorum. Figure 2 plots parts (a) and (b).

**Proposition 3 (Conflict-scaled quorum).** Replace the fixed quorum in A3 by the threshold $k_0 + \gamma \Delta$ of Equation (7b), and let $\Delta_a$ and $\Delta_b$ be the conflicts of the false and the true template with the model's belief before any observation of the situation has been consolidated. Then Proposition 2 holds with $K_a = \lceil (k_0 + \gamma \Delta_a)/\chi_a \rceil$ and $K_h = \lceil (k_0 + \gamma \Delta_b)/\chi_h \rceil$ up to the first consolidation.

*Proof.* Until the first consolidation for the situation, the store contains nothing that matches it, so the conflicts of both templates are fixed by the model's prior and each threshold is a constant. The proof of Proposition 2 applies to each template with its own threshold. ▫

Proposition 3 separates two cases that a fixed quorum treats alike. An attack that contradicts confident knowledge has a large $\Delta_a$ and needs many more credible identities, while learning in a novel situation, where $\Delta_b \approx 0$, keeps the short delay of the base quorum $k_0$. The price is paid by legitimate change: when the world changes and the correct behaviour contradicts what the model confidently believes, $\Delta_b$ is large too and adoption slows. Attacks on situations where the model knows nothing are not made more expensive, so conflict scaling complements rather than replaces credibility and outcome signals.

**Remark (Self-correction requires unforgeable feedback).** Suppose slot $i$ drives the output with gated read weight at least $\acute{\alpha}$ and every such output fails. After $r$ failures $s_i \le s_i(0) e^{-\eta \acute{\alpha} r}$, so the slot falls below a competitor of fixed strength $s'$ after at most $\lceil \log(s_i(0)/s')/(\eta \acute{\alpha}) \rceil$ failures, a number logarithmic in the support with which it was consolidated. The guarantee assumes that feedback on the model's own outputs cannot be forged. If the adversary controls that channel as well, reconsolidation can entrench a false entry.

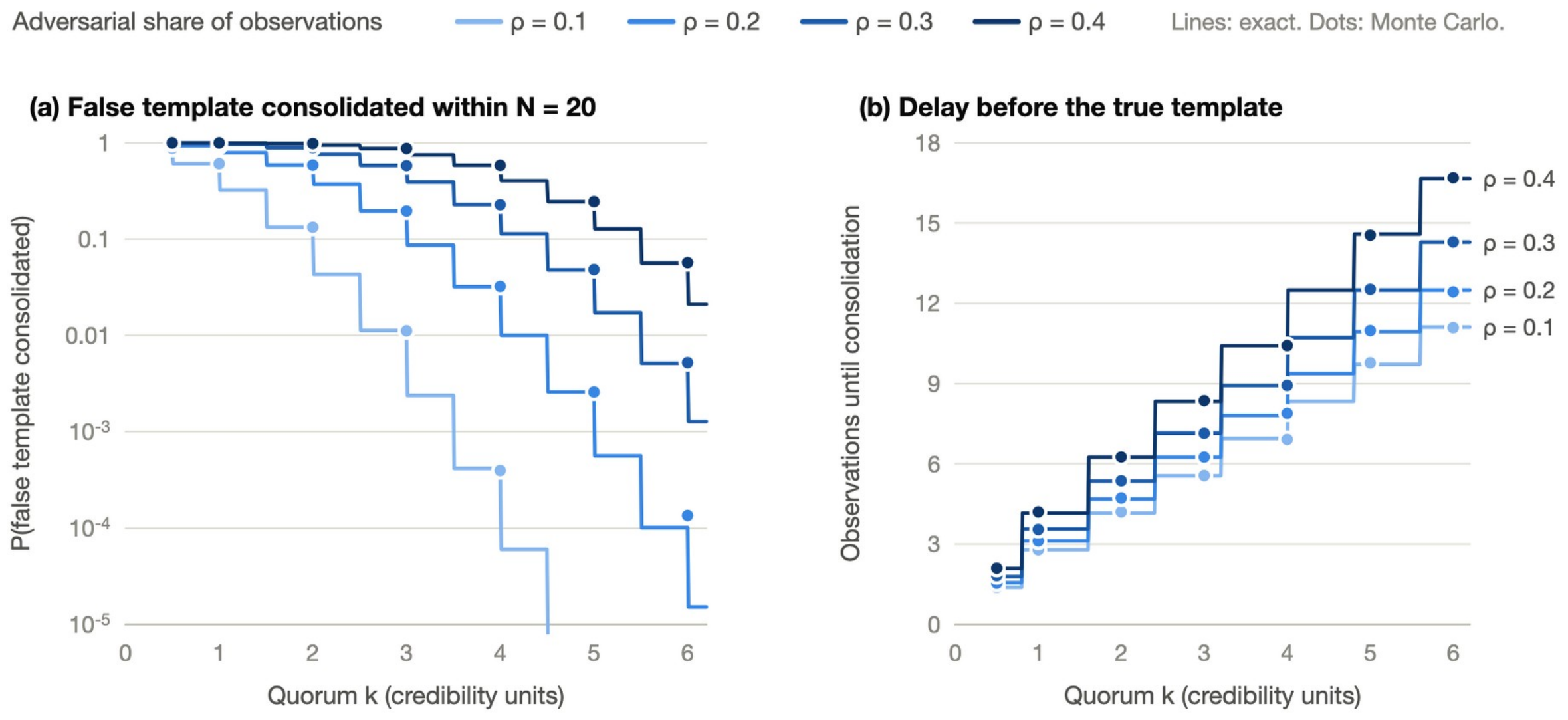

Figure 2. Quorum consolidation under Proposition 2 with $N=20$ observations, $q=0.8$, $\chi_a=0.5$ and $\chi_h=0.8$. Left: probability that a coordinated false template is consolidated within the window (log scale). Right: expected number of observations before the true template is consolidated. Lines show the exact expressions and dots show Monte Carlo estimates.

# 6 Proof-of-concept simulation

## 6.1 Setup

The simulation replaces the language model with an abstract world in which the lifecycle operators can be examined in isolation. The code requires only NumPy and is released with the paper.

*World.* There are 200 situations and 50 behaviour templates. Each situation has a unit key in $\mathbb{R}^{64}$ that is observed with Gaussian noise of norm about 0.35, and each template has a unit embedding. A frozen base model stands in for pretrained knowledge: for a random half of the situations it assigns probability 0.9 to the correct template, and elsewhere it is uniform. At step 3,000 the correct template changes for 20% of the known situations that are not attack targets, so that the base model becomes confidently wrong on them.

*Stream.* Each of 6,000 steps selects a situation uniformly at random. With probability 0.8 the step is an observation, and with probability 0.2 it is an interaction in which the model must produce a template and then receives outcome feedback with probability 0.7. Honest demonstrations come from 100 established sources, whose accuracies are drawn from $Beta(8,2)$ and whose credibility priors reflect a noisy track record, or, for 30% of honest observations, from first-time identities with a $Beta(1,1)$ prior. The outcome of a demonstration is visible with probability 0.7. Twenty situations are attack targets, and 30% of the observations about them are adversarial and show a fixed false template.

*Threat models.* S0 has no attacker. In S1 every injection uses a fresh identity and the adversary cannot fake outcomes, so its demonstrations are seen to fail whenever an outcome is visible. In S2 the adversary still uses fresh identities but reports success for its own demonstrations. In S3 five identities with strong reputations carry all injections and report success. Feedback on the model's own outputs is never forged.

*Memory.* The store holds at most 150 slots and the buffer 400 entries, and support is counted over a window of 1,500 steps. Reads follow Equation (3) with a null slot and a strength prior, and the contribution of memory to the score of a template is the read vector projected onto that template's embedding. Surprise and conflict are computed from the resulting distribution over templates. The constants in Appendix B were set once for all policies and were not tuned per policy or per scenario.

*Policies. v1* appends every observation, evicts the oldest slot and reads without a null slot or strength prior, following the first version. *Surprise-gated* writes an observation when its surprise exceeds 1.5 nats, in the spirit of surprise-driven test-time memories. *Trust-gated* writes when the source's credibility is at least 0.6. *SLS, fixed quorum* applies Equations (7a) and (7b) with $k_0=2$ and $\gamma=0$, and *SLS, conflict-scaled* uses $k_0=1$ and $\gamma=0.25$. All policies except v1 share the corrected read, strength-based eviction, reconsolidation (8) and decay. The ablations remove one component at a time from SLS with a fixed quorum: the absolute quorum ($k_0=0$), the relative condition, confirmations and tallies (which also removes the relative condition), credibility tracking (every source counts 0.7), the payoff signal, reconsolidation, strength-based eviction (replaced by first-in-first-out) and the null slot.

*Metrics.* Every 250 steps the model is evaluated on all 200 situations with fresh key noise and without memory updates. We report accuracy on all situations and on three subsets (known and stable, unknown, and drifted), the attack success rate, defined as the share of target situations answered with the false template, and the mean accuracy on unknown situations over the whole run as a measure of learning speed. Final values average the evaluations after step 5,000. Every configuration runs on the same 20 seeds, so all policies see identical streams, and intervals are 95% t-intervals across seeds.

## 6.2 Results

Table 3. Main results at an adversarial share of $\rho=0.3$. Values are final means over 20 seeds. Attack success is given with 95% confidence half-widths; for the accuracy columns the half-widths are at most 0.045. Learning speed is the accuracy on unknown situations averaged over the whole run. The first version and the surprise-gated policy ignore sources and outcomes, so their results are identical in S1 to S3.

| Policy | S0 accuracy | S0 drifted | S0 learning speed | Entries written (S0) | S1 attack success | S2 attack success | S3 attack success | S3 accuracy |
|---|---|---|---|---|---|---|---|---|
| First version (ungated) | 0.633 | 0.415 | 0.437 | 4,806 | 0.141 ± 0.019 | 0.141 ± 0.019 | 0.141 ± 0.019 | 0.623 |
| Surprise-gated | 0.843 | 0.700 | 0.746 | 1,648 | 0.181 ± 0.024 | 0.181 ± 0.024 | 0.181 ± 0.024 | 0.819 |
| Trust-gated | 0.794 | 0.019 | 0.743 | 318 | 0.013 ± | 0.015 ± | 0.064 ± | 0.780 |

| Policy | S0 accuracy | S0 drifted | S0 learning speed | Entries written (S0) | S1 attack success | S2 attack success | S3 attack success | S3 accuracy |
|---|---|---|---|---|---|---|---|---|
| | | | | | 0.010 | 0.009 | 0.025 | |
| SLS, fixed quorum | 0.996 | 0.958 | 0.853 | 118 | 0.000 ± 0.000 | 0.059 ± 0.021 | 0.066 ± 0.019 | 0.989 |
| SLS, conflict-scaled quorum | 0.992 | 0.910 | 0.914 | 117 | 0.000 ± 0.000 | 0.071 ± 0.028 | 0.075 ± 0.025 | 0.984 |

*Accuracy and efficiency*. Without attacks (S0), both SLS policies end close to perfect accuracy while writing about 118 entries, roughly one fortieth of what the first-version design writes (Table 3). That design suffers on three counts: first-in-first-out eviction discards useful entries, reads without a null slot let irrelevant entries interfere with what the base model already knows (accuracy on known situations 0.927), and honest errors are stored next to correct demonstrations. The surprise-gated policy learns faster but also stores honest errors, which are surprising, and it loses accuracy on known situations (0.916). The trust-gated policy fails almost completely on drifted situations (0.019). It keeps merging confirmations of the old behaviour into one entry, which gives that entry a strength that new demonstrations cannot overcome: a memory that counts frequency without recency entrenches stale knowledge.

*Attacks*. When the adversary cannot fake outcomes (S1), observed failures filter its demonstrations and both SLS policies reduce attack success to zero; without the payoff signal it rises to 0.018. With forged outcomes, attack success for SLS with a fixed quorum is 0.059 (S2) and 0.066 (S3), against 0.141 for the first version and 0.181 for surprise gating. Surprise gating is the most exploitable policy because every attack is surprising by construction. The trust-gated policy reaches lower or similar attack success (0.015 and 0.064), but it pays with the drift failure described above and with lower accuracy (0.78 to 0.79). No policy is best on every measure; at this adversarial share, SLS gives the best balance of accuracy, adaptation and robustness.

Table 4. Ablations of SLS with a fixed quorum. Values are final means over 20 seeds; attack success in S2 and S3 is given with 95% confidence half-widths.

| Variant | S0 drifted | S0 learning speed | S1 attack success | S2 attack success | S2, known targets | S3 attack success | S3, known targets |
|---|---|---|---|---|---|---|---|
| Full lifecycle | 0.958 | 0.853 | 0.000 | 0.059 ± 0.021 | 0.099 | 0.066 ± 0.019 | 0.119 |
| No absolute quorum ($k_0 = 0$) | 0.948 | 0.963 | 0.000 | 0.053 ± 0.016 | 0.073 | 0.028 ± 0.011 | 0.055 |
| No relative condition | 0.967 | 0.853 | 0.000 | 0.110 ± 0.025 | 0.203 | 0.093 ± 0.025 | 0.173 |
| No confirmations or tallies | 0.968 | 0.853 | 0.000 | 0.173 ± 0.028 | 0.177 | 0.146 ± 0.028 | 0.155 |
| Uniform credibility | 0.967 | 0.865 | 0.001 | 0.099 ± 0.026 | 0.162 | 0.058 ± 0.020 | 0.107 |
| No payoff signal | 0.958 | 0.853 | 0.018 | 0.059 ± 0.021 | 0.099 | 0.066 ± 0.019 | 0.119 |
| No reconsolidation | 0.960 | 0.853 | 0.000 | 0.083 ± 0.026 | 0.100 | 0.082 ± 0.027 | 0.141 |
| First-in-first-out eviction | 0.958 | 0.853 | 0.000 | 0.059 ± 0.021 | 0.099 | 0.066 ± 0.019 | 0.119 |
| No null slot | 0.924 | 0.848 | 0.000 | 0.069 ± 0.028 | 0.112 | 0.071 ± 0.022 | 0.127 |

*Ablations*. Table 4 attributes most of the protection to two components. Removing the relative condition raises attack success on known targets from 0.099 to 0.203 in S2 and from 0.119 to 0.173 in S3, and removing confirmations and tallies, which also disables the relative condition, raises overall attack success to 0.173 and 0.146. We added these components after an earlier version of the lifecycle, equivalent to the last of these ablations, failed against forged outcomes. Its surprise gate ignored confirmations of what the model already did while admitting every contradiction, so the adversary accumulated support that honest sources were never credited with. Credibility tracking matters against fresh identities (0.099 without it in S2) but not against reputable ones (0.058 in S3), as expected. Reconsolidation lowers attack success by about 0.02, the null slot mainly helps adaptation to drift (0.958 against 0.924), and the eviction policy has no effect because the store never filled. The absolute quorum is not what protects: removing it gives the lowest attack success in S3 (0.028) and the fastest learning, at the cost of occasional interference from single honest errors (accuracy on known situations 0.998).

*Quorum and conflict scaling.* Raising the fixed quorum from 0 to 6 in S0 lowers the learning-speed measure from 0.963 to 0.241 and accuracy on drifted situations from 0.948 to 0.182 (0.989 at $k_0=1$), the linear delay of Proposition 2(b) compounded over many situations. Under attack, attack success varies little and not monotonically for quorums from 0 to 4 (0.038 to 0.073 in S2), and it falls below 0.01 only at $k_0=6$, where accuracy collapses to 0.64. This is what Proposition 2(d) leads one to expect. Once consolidation is relative, the race between credible honest and adversarial evidence decides the outcome, and the absolute quorum mainly sets how long the contest lasts, which delays recovery from a successful injection as much as the injection itself. Conflict scaling moved learning and adaptation in the directions that Proposition 3 predicts, with faster learning in unknown situations (0.914 against 0.853) and slower adaptation to drift (0.910 against 0.958). Its effect on attacks was small: attack success on known targets was lower (0.079 against 0.099 in S2 and 0.110 against 0.119 in S3, within the confidence intervals) and on unknown targets higher (0.058 against 0.013 in S2).

*Majority reversal.* Figure 4 varies the adversarial share. For SLS with a fixed quorum in S2, attack success rises from 0.003 at $\rho=0.1$ and 0.059 at $\rho=0.3$ to 0.263 at $\rho=0.5$ and 0.744 at $\rho=0.7$, whereas the first version rises more slowly, to 0.354. Once adversarial evidence outpaces honest evidence, relative consolidation adopts the false template more reliably than an ungated memory that averages whatever it has recently seen. Proposition 2(c) and (d) predict this reversal: conformity amplifies whichever side is faster. In S3 the trust-gated policy fares worst at $\rho=0.7$ (0.581), because reputable identities pass its threshold. Figure 3 shows the dynamics behind these results.

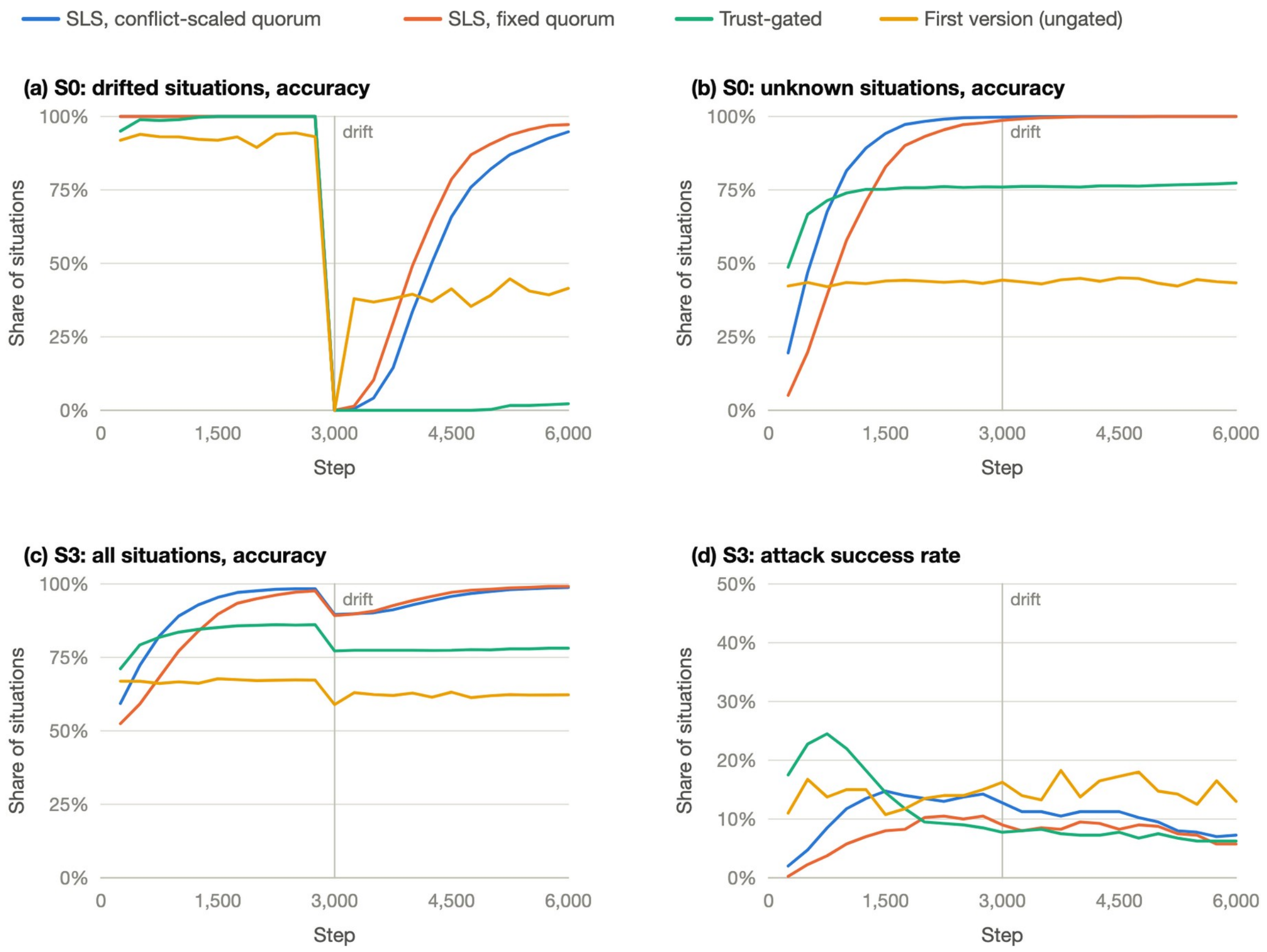

Figure 3. Simulation dynamics averaged over 20 seeds. (a, b) Without attacks: accuracy on drifted situations, whose correct behaviour changes at step 3,000, and on situations unknown to the base model. (c, d) Under reputable identities with forged outcomes (S3): accuracy on all situations and attack success on the 20 target situations.

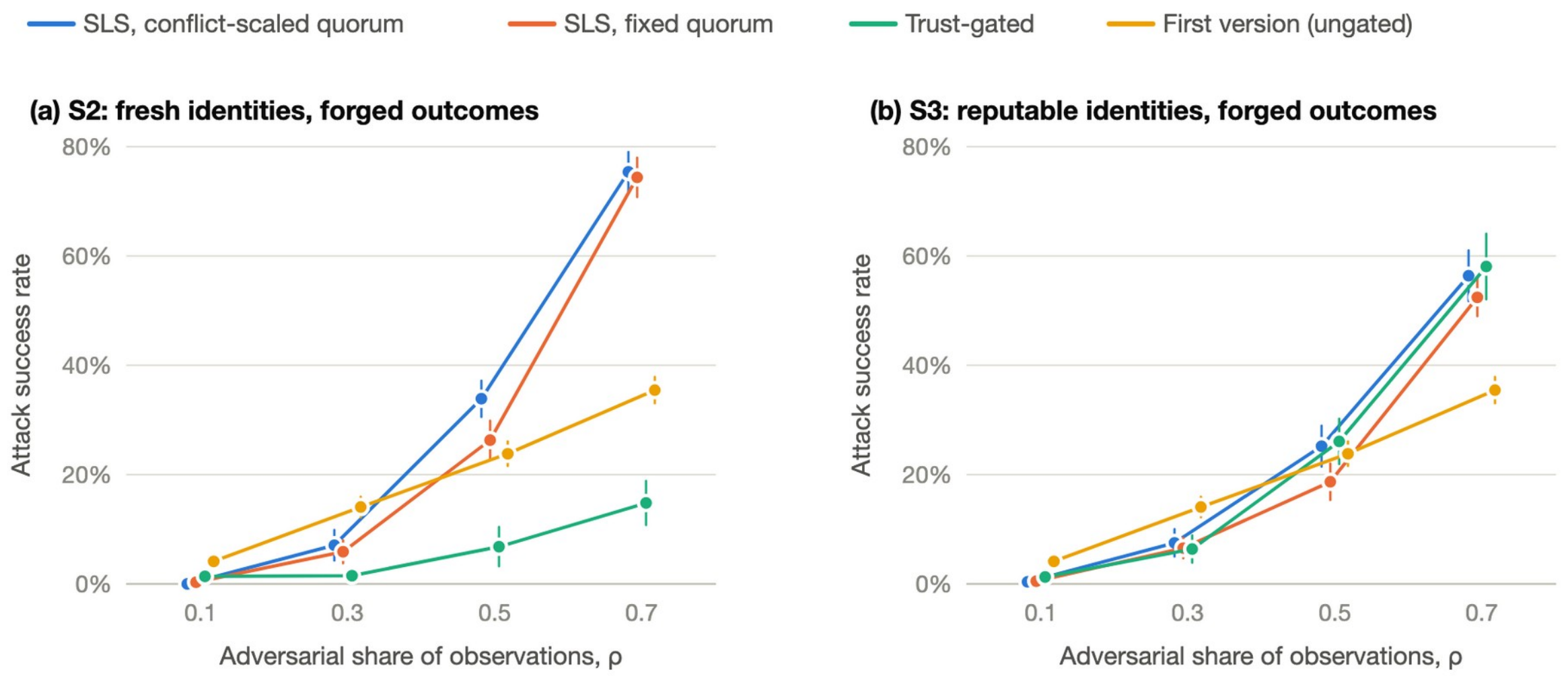

Figure 4. Attack success at the end of the run as the adversarial share of observations about target situations varies, for fresh identities (S2) and reputable identities (S3), both with forged outcomes. Dots are means over 20 seeds and vertical lines are 95% confidence intervals.

### 6.3 What the simulation shows and what it does not

The simulation shows that the operators specified in Section 4 behave as intended when their inputs are available. Selective writing avoids the interference and wasted capacity of the first version, outcome and credibility signals filter demonstrators, relative consolidation protects while credible honest evidence outpaces adversarial evidence, larger quorums cost the delay that Proposition 2 predicts, and conflict scaling trades faster learning for slower adaptation as Proposition 3 predicts. It also shows where the design fails: once adversarial evidence arrives faster than honest evidence, relative consolidation adopts the false template more reliably than a memory without it, and conflict scaling does little for situations the model knows nothing about.

The simulation does not show that a language model can compute these signals reliably, that the read path can be trained to use such a store, or that the benefits survive the noise of real identities and outcomes. Its situations are clean clusters, its templates are discrete, its base model is a fixed table, and its sources and outcomes are given rather than inferred. Section 7 describes the experiments needed to close these gaps.

## 7 Evaluation protocol for language-model instantiations

*Model.* Start from an open-weights decoder of moderate size, insert Retention reads in a subset of blocks, for example every fourth block, with zero-initialised output projections (Proposition 1), and keep the base weights frozen in a first study.

*Tasks.* Three families of tasks are needed. Multi-session conversation benchmarks test whether facts and preferences survive session boundaries (Xu et al. 2022; Maharana et al. 2024; Wu et al. 2025). Long-context reasoning benchmarks check that memory reads do not degrade the use of the context itself (Kuratov et al. 2024). The third family does not yet exist, and it is the one this paper motivates: *observational procedure learning*. Several simulated users demonstrate a procedure across sessions, such as a formatting convention, a tool-use workflow or a correction of a recurring error, with known identities and outcome signals, and later sessions require the model to reproduce the procedure without the demonstration in context. Adversarial variants inject false procedures following published threat models for memory poisoning (Chen et al. 2024; Dong et al. 2025), with fresh identities, reputable identities, forged outcomes, and conflicts with pretrained knowledge.

*Baselines.* No memory; the full history in context where it fits; retrieval over past sessions; memories with forgetting curves and strengthening on recall (Zhong et al. 2024); latent memory pools (Wang et al. 2024, 2025); and surprise-only and trust-only write policies implemented on the same Retention Layer.

*Measurements.* Cross-session task accuracy; adoption delay, measured in sessions, before a demonstrated procedure is reproduced; interference with pretrained knowledge on held-out tasks; attack success rate for each threat model; store growth and inference overhead; and deletion fidelity, meaning whether removing a source's contributions also removes their effect on behaviour (Cao and Yang 2015). Results should be reported over several seeds with confidence intervals, and quorum and conflict-scaling sweeps should be reported in full, because Proposition 2 predicts a trade-off rather than a single best setting.

*System-level replay.* A deployed assistant that already logs its conversations allows a cheaper first test than training reads into a model. The recorded stream of turns can be replayed through alternative write, consolidation and forgetting policies with the logged responses held fixed. Each policy is then scored by whether it still holds, and ranks highly, the entries that the user later refers back to, identified by annotation or by an automatic judge whose agreement with annotators is reported. Synthetic observations injected into the replayed stream show what each policy keeps from a simulated adversary. Replay cannot show that memory changes behaviour, which requires an online comparison, but it lets the choice of lifecycle be studied separately from the training of the read path. Such logs contain personal data, so replay studies need consent and should report only aggregate results.

# 8 Applications and deployment considerations

## 8.1 Application domains

The first version listed six application domains. The social learning view changes what each of them requires, because it asks who the sources are, which outcomes are observable and what an adversary can forge.

*Personal assistants.* In a per-user store the user is usually the only relevant source, so a quorum across distinct sources is meaningless and a single credible confirmation should suffice. Shared memories that learn from many users, for example conventions common across a team, are where quorum consolidation applies. The two stores need different parameters, and personal content should never be consolidated into shared memory without consent. Section 8.2 reports what a long-running single-user assistant shows about such stores.

*Fraud detection.* Analysts' decisions are demonstrations, chargebacks are delayed outcomes, and fraudsters are adversaries who probe the system. Conflict scaling suits this setting: established rules require broad agreement before they are overturned, while a new pattern reported by several analysts can be adopted quickly.

*Content moderation.* Moderator decisions and appeal results supply demonstrations and outcomes. Coordinated reporting campaigns are an instance of the majority condition in Proposition 2(c), so counts of reports must be weighted by credibility rather than taken at face value.

*Robotics.* Operator demonstrations with task success signals fit the observation format directly. Because errors have physical consequences, consolidated behaviours should be validated before they are executed autonomously.

*Clinical decision support.* The first version suggested that a system could retain each new patient case and refine its diagnostic behaviour. That claim needs qualification. Patient data are regulated, outcomes are delayed and confounded, and consolidated patterns should be reviewed by clinicians before they influence care. The provenance kept in support sets makes such review possible, but this paper makes no claim of clinical benefit.

### 8.2 Lessons from a long-running assistant

The lifecycle in Section 4 has not yet been built into the blocks of a language model. A system-level analogue of parts of it has been in use since November 2025 in the Hybrid Cognitive System (HCS), a single-user conversational assistant that the author built around a language model. Its design has changed substantially since then, and the account below describes the system as documented in September 2026. HCS keeps past turns in a store outside the model. It scores each entry for importance from the emotion, intensity, topic and repetition of the turn, and it ages entries on schedules that lengthen with importance. Retrieval fuses a character n-gram TF-IDF ranking with a BM25 ranking by reciprocal rank fusion, reweights the result by importance and recency, and writes the retrieved entries into the prompt before every response. Two ledgers record delayed outcomes. One notes whether a joke by the assistant landed, failed or was ignored, judging by the user's next message, and updates Beta-Bernoulli posteriors that set how much humour the assistant uses. The other scores the assistant's advice on tone by the valence of the next message.

HCS does not implement the strategies of Section 2.2. With a single user it has no quorum across sources and no model of source credibility, and the effect of its memory on the model's responses has not been measured in a controlled comparison. Its operating record is therefore not evidence for the claims of this paper. It is informative because several problems that the lifecycle addresses appeared in HCS without being sought, and the fixes adopted there resemble the corresponding operators.

*Frequency without recency entrenches.* HCS merges repeated statements of a fact into one entry. If a merge could only raise confidence, a fact that had stopped being true could never be displaced, so facts lose confidence with a half-life of 14 days and single-valued predicates carry a flag that lets a newer value supersede an older one. The trust-gated policy in Section 6 failed on drifted situations for the same reason, and windowed support and decay are the lifecycle's response.

*A model's own output is not independent evidence.* HCS stores its replies next to the user's messages. The replies carried more emotional language and emoji, so their importance scores were higher (median about 0.56, against about 0.30 for the user's messages), and they outranked the user's own words in a score that also enters retrieval. Replies are now down-weighted by a factor of 0.7. In the terms of this paper the model is one more source, and its agreement with itself says the least about the world. A lifecycle that reads its outputs back should not count them toward support or treat them as confirmations.

*Duplicated events are not agreement.* The pipeline sometimes submitted the same turn more than once. In one recorded case a single message produced four submissions, and each opened its own memory entry, personality event and inference record. Turns with the same role and normalised text within ten minutes are now merged. Support in Section 4.4 counts distinct sources within a window for the same reason.

*Outcomes arrive late and are ambiguous.* Reconsolidation, Equation (8), assumes that the outcome of a reproduced behaviour is observed. In HCS a laugh within three minutes of a joke counts as success, weighted by how strong the laugh is. A cold or negative reply within three minutes counts as failure, and a reply after more than ten minutes, or none at all, counts as weak negative evidence with weight 0.4. Any instantiation of the lifecycle has to make the same choices explicit: how long to wait for an outcome, how to weight partial success and how to treat silence.

*Some content must be kept after a single mention.* HCS compresses episodic records into summaries after about 30 days and ignores words that occur fewer than twice, so a hotel name, an appointment or a promise mentioned once disappears when the raw record is pruned. A quorum has the same blind spot. When users state facts about themselves or make commitments, the speaker matters more than the count, because such statements are testimony from the principal rather than observations of other people's behaviour. Developmental research draws a similar line. Children learn about things they cannot observe from testimony, prefer informants who have been accurate before and treat demonstrations addressed to them as conveying generalisable knowledge (Harris and Koenig 2006; Koenig and Harris 2005; Csibra and Gergely 2009). People also check both the competence and benevolence of a source and the coherence of its message with what they already believe (Sperber et al. 2010), which are the roles of credibility and conflict in Sections 4.4 and 4.5. A deployed Retention Layer should record principal testimony on a single occurrence, as HCS does for reminders and stated intentions, and reserve quorum consolidation for generalisations drawn from others.

*A memory can fail silently.* Proposition 1 guarantees that a block whose store is empty, or whose gate is closed, computes the base model exactly. The guarantee has a cost that HCS makes visible: from the outside, a memory that has stopped working looks like a memory with nothing relevant to return. In HCS an invalid interpreter path is enough to stop every stored entry from reaching the prompt while the assistant keeps answering normally. The background loop that synthesises memories has not completed a cycle since 27 May 2026, a failure that raised no error and was found by a health check. That check now tests whether each module still writes its state and whether retrieved entries still reach the prompt. A deployed Retention Layer needs an equivalent check, for example an alert when the share of reads dominated by the null slot rises.

These observations come from one system and one user, and none of them measures a benefit. They point to three refinements that the simulation in Section 6 could not reveal, because it had no self-generated content, no testimony and no infrastructure that could fail: exclude a model's own outputs from support, record principal testimony outside the quorum, and monitor the read path. The logs of such a system also permit the replay evaluation described in Section 7.

## 9 Limitations

The theoretical results rest on independence assumptions. Honest errors that coincide, such as a misconception shared by many users, violate A2, and a quorum will then consolidate the shared error; conformist transmission has the same weakness in human groups (Boyd and Richerson 1985). Credibilities in Proposition 2 are constant, whereas in practice they evolve and can be manipulated over time.

The simulation is small and abstract, and its constants were chosen by hand. The first-version baseline is our reconstruction of an underspecified design, and the surprise-gated and trust-gated policies are simplified stand-ins for families of methods, not reimplementations of particular systems. Confirmations, tallies and the relative condition were added after an earlier variant of the lifecycle failed in these simulations. The change follows from what conformist transmission requires, and the earlier variant is reported as an ablation, but the revised lifecycle has not yet been tested on scenarios designed independently of that observation.

Conflict scaling makes the model conservative. Legitimate changes that contradict confident knowledge are adopted more slowly, and the size of that delay depends on how well the model's confidence is calibrated.

The architecture assumes that source identities and outcome signals exist. Many deployments have neither, and inferring them from text opens further attack surfaces. Stored entries also become stale when base weights change, and per-block stores add memory and latency that we have not measured at scale.

The operators in Section 4 treat everything they receive as observations of behaviour. Statements that users make about themselves, and commitments made to them, are testimony that should be kept after one occurrence (Section 8.2), and the lifecycle does not yet distinguish the two.

## 10 Ethics, privacy and governance

*Consent and separation.* Personal memories and shared memories should be stored separately, and contributions should move from the first to the second only with consent.

*Deletion.* Support sets record which sources contributed to each entry. Entries supported only by a departing user can be deleted; for merged entries, that user's credibility can be subtracted from the support and the quorum checked again. Memories written into weights do not allow this without retraining, which is one reason to prefer an explicit store where the right to erasure applies (European Parliament and Council of the European Union 2016; Cao and Yang 2015).

*Fairness.* Credibility priors favour established identities, and conformity discounts minority practices. A memory governed by these strategies can therefore reproduce existing inequalities, for example by giving less weight to newcomers or to users whose practices differ from the majority's. Mitigations include domain-specific credibility, caps on credibility, keeping personal preferences in per-user stores and auditing consolidation rates by group.

*Manipulation and misuse.* The mechanisms that resist poisoning could also be configured to make a model resist legitimate correction. Deployments should disclose what is retained, from whom and for how long, and should let users inspect and remove what the system has learned from them.

## 11 Conclusion

The first version of this paper asked Transformers to retain what they observe, and it borrowed the vocabulary of social learning theory to describe how. This revision takes the borrowing seriously. A deployed model learns mostly from others, so its memory needs rules for when to trust observed behaviour, whom to trust and how much agreement to demand, and social learning strategies supply such rules together with their known failure modes. Built into a corrected Retention Layer, these rules yield a memory that reduces exactly to the base model when empty, admits observations by surprise, outcome and credibility, consolidates them by a quorum scaled to conflict, and revises them by the consequences of reproduction. The analysis quantifies the central trade-off between robustness and delay and states when conformity becomes a liability, and the simulation shows the predicted behaviour in a controlled setting. Whether the approach holds up inside real language models, with real identities and noisy outcomes, is the question that the protocol in Section 7 is designed to answer.

## Declarations

*Use of AI tools.* The first version acknowledged the use of Grammarly, DeepL and ChatGPT for language editing. For this revision an AI assistant (Claude, Anthropic) supported literature search, reference verification, drafting of text, the simulation code and the figures. The author reviewed all content and takes responsibility for it.

*Competing interests.* The author develops the Hybrid Cognitive System described in Section 8.2 and may use it commercially. The author declares no other competing interests.

*Code availability.* The simulation code (`sls_retention_sim.py`), the figure script and the raw results (`results.json`) are available at https://github.com/myaslioglu/retention-layer under the GNU General Public License v3.0. Version 1.0.0, which produced the results reported here, is archived on Zenodo (Yaslioglu 2026).

## Appendix A. Proof of Proposition 2(c)

Write $r_a=\rho$ and $r_h=(1-\rho)q$. The waiting time $T_a$ for the $K_a$-th adversarial observation is negative binomial with mean $\mu_a=K_a/r_a$ and variance $K_a(1-r_a)/r_a^2$, and $T_b$ is negative binomial with mean $\mu_b=K_h/r_h$ and variance $K_h(1-r_h)/r_h^2$. Since $K_a=k/\chi_a+O(1)$ and $K_h=k/\chi_h+O(1)$,

$$\mu_a=\frac{k}{\chi_a\rho}+O(1),\mu_b=\frac{k}{\chi_h(1-\rho)q}+O(1),Var(T_a),Var(T_b)=O(k).$$

Suppose $\chi_a\rho>\chi_h(1-\rho)q$. Then $\mu_b-\mu_a=ck+O(1)$ for some constant $c>0$. Let $\acute{\mu}=(\mu_a+\mu_b)/2$. If $T_a<\acute{\mu}<T_b$, then $T_a<T_b$, so

$$Pr[T_a\geq T_b]\leq Pr[T_a\geq\acute{\mu}]+Pr[T_b\leq\acute{\mu}]\leq\frac{Var(T_a)}{(\acute{\mu}-\mu_a)^2}+\frac{Var(T_b)}{(\mu_b-\acute{\mu})^2}=\frac{O(k)}{\Theta(k^2)}\to 0$$

by Chebyshev’s inequality, whatever the dependence between $T_a$ and $T_b$. The same observation cannot count toward both templates, so $Pr[T_a=T_b]=0$, and therefore $Pr[T_a<T_b]\to 1$. The case $\chi_a\rho<\chi_h(1-\rho)q$ is symmetric. ▫

**Part (d).** Let $A_n$ and $H_n$ be the adversarial and honest supports after $n$ observations and $D_n=A_n-H_n$. $D_n$ is a random walk whose increments $X$ equal $\chi_a$ with probability $\rho$, $-\chi_h$ with probability $(1-\rho)q$, and $0$ otherwise. The false template can be consolidated at step $n$ only if $A_n\geq k$ and $A_n$ exceeds the support of every rival, in particular $D_n>0$; hence only at some step $n\geq T_a$ with $D_n>0$.

Suppose $E[X]=\chi_a\rho-\chi_h(1-\rho)q<0$. Because $X$ is bounded and positive with positive probability, $\phi(\theta)=E[e^{\theta X}]$ is finite, $\phi(0)=1$, $\phi'(0)<0$ and $\phi(\theta)\to\infty$ as $\theta\to\infty$, so there is $\theta^*>0$ with $\phi(\theta^*)=1$. By Lundberg’s inequality, a walk with these increments that starts at $-u$ ever reaches $[0,\infty)$ with probability at most $e^{-\theta^* u}$. Applying the strong Markov property at $T_a$,

$$Pr[a \text{ is consolidated}] \le Pr\left[D_{T_a} > -c\,k\right] + e^{-\theta^* c\,k},\; c = \frac{1}{2}\left(\frac{\chi_h(1-\rho)q}{\chi_a \rho} - 1\right) > 0\,.$$

When the adversarial support first reaches $k$, about $k/(\chi_a \rho)$ observations have been made and the honest support is about $k\chi_h(1-\rho)q/(\chi_a\rho)$, so $D_{T_a}$ is close to $-2ck$. Chernoff bounds for the binomial counts over a number of observations proportional to $k$ make $Pr[D_{T_a} > -c\,k]$ exponentially small in $k$. If instead $E[X] > 0$, then $D_n \to \infty$ and $A_n \to \infty$ almost surely, so both conditions eventually hold. ▫

## Appendix B. Simulation constants

Table B1. Constants of the simulation in Section 6. The code that produced every number in Section 6 is released with the paper, and the numerical check of Proposition 1 uses a random block with $d=16$, four heads, seven tokens and five memory slots.

| Quantity | Value |
|---|---|
| Key dimension; situations; templates | 64; 200; 50 |
| Situations known to the base model; its confidence | 50%; probability 0.9 on the correct template |
| Observation noise on keys | Gaussian, norm about 0.35 |
| Steps; share of interaction steps; feedback probability | 6,000; 0.2; 0.7 |
| Drift | 20% of known, non-target situations change at step 3,000 |
| Established honest sources; their accuracy | 100; drawn from $Beta(8,2)$ |
| Share of honest observations from first-time identities | 30% |
| Credibility prior of established sources | $Beta(1+4\hat{q}, 1+4(1-\hat{q}))$, with $\hat{q}$ the accuracy plus $N(0,0.1)$ noise, clipped to $[0.05, 0.95]$ |
| Credibility prior of first-time identities; of reputable attackers (S3) | $Beta(1,1)$; $Beta(4.6,1.4)$ |
| Attack targets; adversarial share $\rho$; outcome visibility | 20; 0.3; 0.7 |
| Store capacity; buffer capacity; support window $W$ | 150 slots; 400 entries; 1,500 steps |
| Key temperature; null-slot logit $b_0$; strength-prior weight $\beta$ | 0.10; 6.0; 1.0 |
| Scale of the memory read in template scores | 10 |
| Similarity threshold $\delta$ | 0.70 |
| Reference surprise $\upsilon_0$; gate weights $\theta_\upsilon, \theta_\pi, \theta_\chi$ | 1.5 nats; 2, 2, 2 |
| Reconsolidation step $\eta$; strength decay $\lambda$ per step | 0.5; $2 \times 10^{-4}$ |
| Quorum of SLS, fixed; conflict-scaled | $k_0=2, \gamma=0$; $k_0=1, \gamma=0.25$ |
| Threshold of the trust-gated policy; of the surprise-gated policy | credibility 0.6; surprise 1.5 nats |
| Credibility used in the uniform-credibility ablation | 0.7 |
| Evaluation | every 250 steps on all situations with fresh key noise; final values average steps 5,250 to 6,000 |